\pdfoutput=1
\documentclass[11pt]{article}

\usepackage{acl}
\usepackage{times}
\usepackage{latexsym}

\usepackage[T1]{fontenc}
\usepackage[utf8]{inputenc}

\usepackage{microtype}

\usepackage{inconsolata}

\usepackage{subfigure}
\usepackage{xspace}
\usepackage{graphicx}
\usepackage{tcolorbox} % color box for text
\usepackage{textcomp}
\usepackage{booktabs}
\usepackage{amsmath}
\usepackage{multirow}
\usepackage{multicol}
\usepackage{CJKutf8}
\usepackage[ruled,vlined]{algorithm2e}
\DeclareMathOperator*{\argmax}{argmax}
\usepackage{url}

\def\figref#1{Figure~\ref{fig:#1}}
\def\figlabel#1{\label{fig:#1}\label{p:#1}}

\def\tabref#1{Table~\ref{tab:#1}}

\def\tablabel#1{\label{tab:#1}\label{p:#1}}

\def\secref#1{\S\ref{sec:#1}}
\def\seclabel#1{\label{sec:#1}}

\def\eqlabel#1{\label{eqn:#1}}

\def\appref#1{Appendix~\ref{app:#1}}
\def\applabel#1{\label{app:#1}\label{p:#1}}

\title{Decomposed Prompting: Probing Multilingual Linguistic Structure Knowledge in Large Language Models}

\author{
Ercong Nie\textsuperscript{1,2},~
Shuzhou Yuan\textsuperscript{3},~
Bolei Ma\textsuperscript{2,4},~
Helmut Schmid\textsuperscript{1},\vspace{3.5pt}\\
\textbf{
Michael Färber\textsuperscript{3},~
Frauke Kreuter\textsuperscript{2,4,5}
~and 
Hinrich Schütze\textsuperscript{1,2}\vspace{3.5pt}}
\smallskip
\\
\textsuperscript{1}Center for Information and Language Processing (CIS), LMU Munich, \\
\textsuperscript{2}Munich Center for Machine Learning (MCML), \textsuperscript{3}ScaDS.AI and  TU Dresden,\\
\textsuperscript{4}Department of Statistics, LMU Munich,~\textsuperscript{5}University of Maryland, College Park \smallskip\vspace{3.5pt}
\\
\smallskip
\texttt{nie@cis.lmu.de,~shuzhou.yuan@tu-dresden.de,~bolei.ma@lmu.de,}
}

\begin{document}
\begin{CJK}{UTF8}{gbsn}
\maketitle
\begin{abstract}
Probing the multilingual knowledge of linguistic structure in LLMs, often characterized as sequence labeling, faces challenges with maintaining output templates in current text-to-text prompting strategies. 
To solve this, we introduce a \textit{decomposed prompting} approach for sequence labeling tasks.
Diverging from the single text-to-text prompt, our prompt method 
generates for each token of the input sentence an individual prompt which asks for its linguistic label.
We test our method on the Universal Dependencies part-of-speech tagging dataset for 38 languages, using both English-centric and multilingual LLMs. Our findings show that \textit{decomposed prompting} surpasses the \textit{iterative prompting} baseline in efficacy and efficiency under zero- and few-shot settings. 
% Further analysis reveals the influence of evaluation methods and the use of instructions in prompts. 
% Our multilingual investigation shows that English-centric language models perform better on average than multilingual models.
Moreover, our analysis of multilingual performance of English-centric LLMs yields insights into the transferability of linguistic knowledge via multilingual prompting. 
% Our study offers insights into the multilingual transferability of English-centric LLMs, contributing to the understanding of their multilingual linguistic knowledge.
\end{abstract}

\section{Introduction}
Current Large Language Models (LLMs), such as GPT-3, GPT-4, PaLM, and LLaMA~\citep{brown2020language, chowdhery2023palm, touvron2023llama}, have demonstrated remarkable capabilities in in-context learning across a broad spectrum of language understanding and generation tasks~\citep{zhao2023survey, zhang2023instruction, ziyu-etal-2023-lens}. These models are predominantly trained on massive amounts of English text data, with some limited exposure to other languages. For instance,  LLaMA2's pretraining corpus comprises over 89\% English content, with the rest in other languages or code~\citep{touvron2023llama2}. Yet, these English-centric LLMs~\footnote{In this paper, we regard a model pretrained primarily on English text as English-centric.} exhibit effective performance in complex multilingual language understanding tasks~\citep{deng2023multilingual, wang2023all}.  In multilingual evaluation with prompting, a model performs tasks by directly generating outputs based on a task description and/or a few examples provided in a pivot language (typically English), along with input in a different target language~\citep{ahuja-etal-2023-mega}.
Despite the remarkable multilingual performance of LLMs, the extent and nature of their cross-lingual capabilities remain underexplored~\citep{ye2023language}. 

% This raises a critical question: \textit{Does the multilinguality of these models stem from a deep, generalizable multilingual linguistic understanding, or merely from the superficial alignment of lexical patterns across languages? }

\begin{figure}[!t]
	\centering  
        \includegraphics[width=\linewidth]{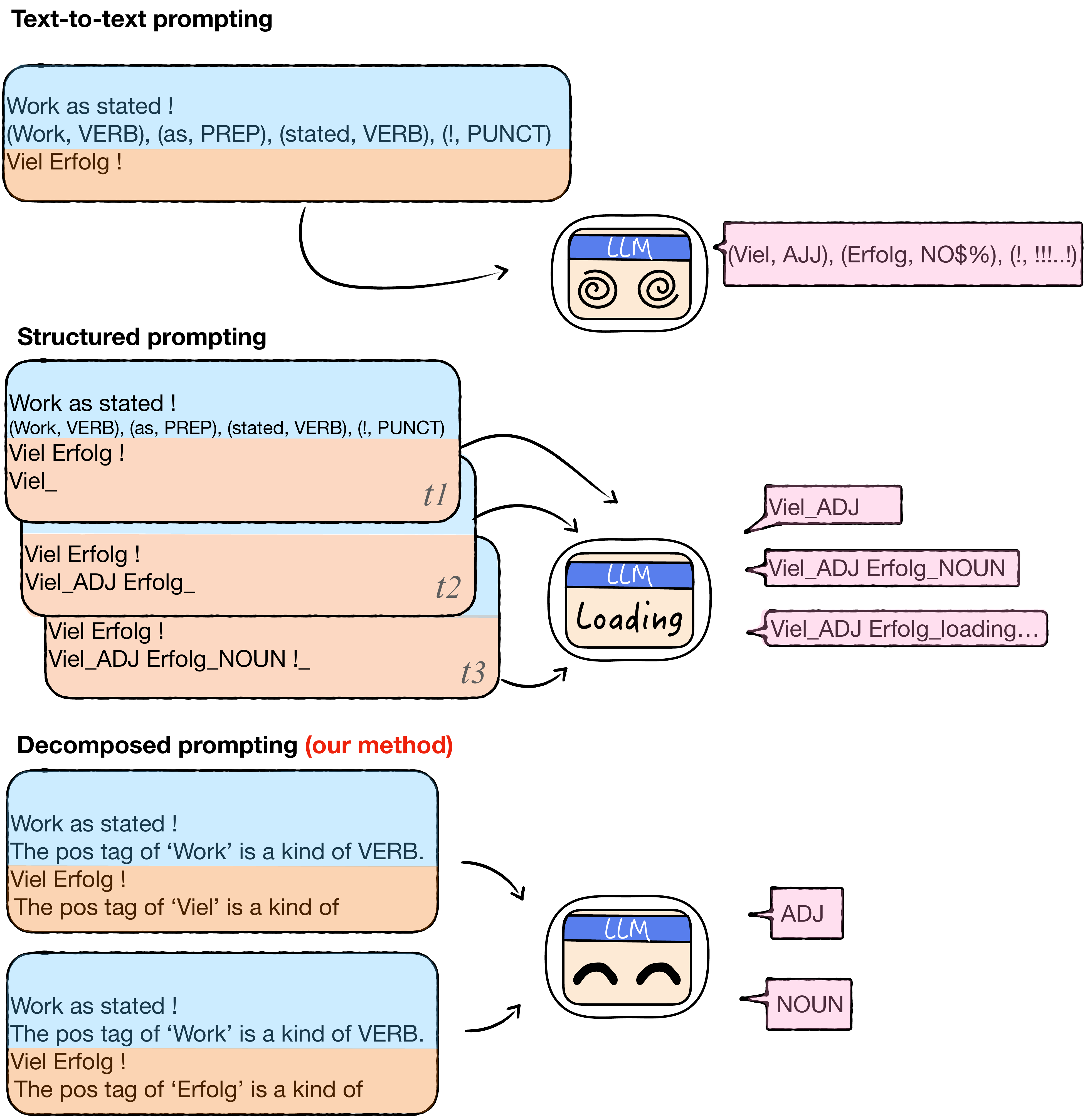}
	\caption{Comparison of different prompting methods for sequence labeling.}

	\figlabel{fig1}   

\end{figure}
 We hypothesize that these models harbor substantial multilingual knowledge. This knowledge, particularly relating to linguistic structure, is commonly conceptualized through sequence tagging tasks~\citep{jurafsky2000speech}. 
 However, the current prompting strategies designed for sequence labeling in LLMs are not well suited to test. For instance, behavioral probing methods~\citep{belinkov-etal-2020-interpretability}, aimed at measuring knowledge stored in language models, struggle to adapt to tasks predicting more complex structures. To overcome the challenges in probing the multilingual knowledge of linguistic structure in LLMs characterized as sequence labeling, drawing inspiration from the token-level prompt-based fine-tuning method by~\citet{ma2024topro}, we introduce the \textit{\textbf{decomposed prompting}} strategy, aiming to probe English-centric LLMs for their understanding of linguistic structure framed as sequence labeling tasks. 
As shown in \figref{fig1}, instead of employing a single text-to-text prompt for labeling an entire sequence in one step, our method decomposes this process into multiple discrete prompts.
More precisely, we first split the input sentence into tokens. Subsequently, we generate an individual prompt for each token which inquires about its linguistic label.

We evaluate our approach on the Universal Dependency (UD) part-of-speech (POS) tagging dataset~\citep{nivre-etal-2020-universal} covering 38 languages with 3 English-centric LLMs and 2 multilingual LLMs. Our approach outperforms the iterative prompting baseline in both zero- and few-shot settings in terms of accuracy and efficiency. 
% We investigate the nuanced impact of evaluation methods and the usage of task instructions within prompts on the performance of decomposed prompting, followed by an empirical comparative study of decomposed and iterative prompting. 
Furthermore, our investigation into the multilingual performance of English-centric LLMs offers valuable insights into their capabilities of transferring linguistic knowledge through multilingual prompting.

% Our contributions are summarized: ...

\section{Background and Related Work}
\seclabel{background}

\paragraph{Multilinguality of English-Centric LLMs}
English-centric LLMs are primarily pretrained on large English text data, with a limited exposion to multilingual data. LLaMA~\citep{touvron2023llama}, for example, is pretrained on an extensive scale of corpora comprising over 1.4 trillion tokens, of which less than 4.5\% constitute multilingual data from 20 different languages. LLaMA 2~\citep{touvron2023llama2} expands this linguistic diversity, featuring 27 languages each representing more than 0.005\% of the pertaining data. 
% Notably, a minor fraction (approximately 4.5\%) of LLaMA's pertaining dataset consists of code data, which, being non-natural language data, falls outside the scope of this paper's discussion.
Mistral 7B~\citep{jiang2023mistral} achieves superior performance and efficiency through the adoption of advanced attention techniques such as Sliding Window Attention (SWA)~\citep{child2019generating}, facilitating faster inference.
% Moreover, the tokenizer of English-centric LLMs is often engineered to support byte-level encoding~\citep{workshop2022bloom, zhang2022opt, touvron2023llama}. This capability allows for the processing of a wide range of scripts beyond the Latin alphabet. 
To enhance the robustness of multilingual processing, the Byte-level Byte-Pair-Encoding (BBPE) algorithm~\citep{sennrich-etal-2016-neural, wang2020neural} is commonly used for tokenization in LLMs.
This approach is able to decompose UTF-8 characters, which are outside the scope of the model vocabulary, into their constituent bytes. Thus, BBPE tokenization equips LLMs with the versatility to handle scripts from any language, theoretically, even those not encountered during training.
In summary, \textit{limited exposure to non-English data} and \textit{byte-level encoding capability}, these two factors discussed above, jointly contribute to the robust multilingual abilities observed in English-centric LLMs.

% \paragraph{Multilingual LLMs and Instruction Tuning}
% LLMs are typically instruction-tuned models, including multilingual LLMs, as exemplified by models such as BLOOMZ~\citep{muennighoff-etal-2023-crosslingual}, which is derived from BLOOM~\citep{workshop2022bloom}, and mTk~\citep{wang-etal-2022-super} which is based on mT5~\citep{xue-etal-2021-mt5}.
% Instruction tuning is a widely utilized method to improve the task understanding and interactivity of LLMs~\citep{zhang2023instruction}. 
% Recent studies augmented the multilingual capabilities of LLMs by applying the multilingual instruction tuning~\citep{kew2023turning, chen2023monolingual, shaham2024multilingual}.  Multilingual LLMs have also been tailored for specific language clusters, such as the SeaLLMs for Southeast Asian Languages~\citep{nguyen2023seallms}. In our work, we utilize instruction-tuned English-centric and multilingual LLMs for our experiments.

\paragraph{Prompting for Sequence Labeling}
% Prompting methods have seldom been applied to sequence labeling tasks. However, some studies have used prompt-based fine-tuning for this purpose. \citet{cui-etal-2021-template} have implemented template-based prompting techniques with the BART model for Named Entity Recognition (NER) tasks. \citet{ma-etal-2022-template} introduced a template-free prompting strategy for few-shot NER, termed entity-oriented LM fine-tuning. Similar to our approach, \citet{ma2024topro} utilized a decomposition-based prompting method to fine-tune multilingual encoder models for cross-lingual sequence labeling. Contrary to these studies, which all operate within a fine-tuning framework, our research is distinctively focused on facilitating multilingual in-context learning in LLMs without the necessity for parameter updates. 

Prompting LLMs for sequence labeling tasks remains a challenge~\citep{ahuja-etal-2023-mega}. 
While \textit{text-to-text prompting} is widely adopted across various benchmarking tasks for LLMs~\citep{lai-etal-2023-chatgpt}, their application to sequence labeling is hindered by the challenges in maintaining the output templates~\citep{asai2023buffet}. 
In response, a decent \textit{iterative prompting} strategy for structured prediction has been introduced~\citep{blevins-etal-2023-prompting} (\figref{fig1}).
In this approach, the model decodes in step $t_i$ a label for the word at position $t_i$ of the sequence. This predicted label, along with the next word, is then input back into the model to predict the next label. However, the dependency of each token’s prediction on the preceding one substantially slows down the inference process. In contrast, our proposed \textit{decomposed promptin}g method offers improvements in both efficacy and efficiency.
Our method is similar to~\citet{ma2024topro} in that both methods decompose an input sentence into a series of prompts; however, their method is used for fine-tuning, while our method is in an in-context learning paradigm without training.

\section{Decomposed Prompting for LLMs}\seclabel{method}
% In this study, we introduce a novel approach for conducting sequence labeling with LLMs through in-context learning, termed \textit{decomposed prompting}. 

% \subsection{Problem Formulation}
Given a test sequence set $\mathcal{X}_{test}$, a label set $L$, and an LLM $M$, we approach the task of sequence labeling as follows: for an input sequence $X \in \mathcal{X}_{test}$ of length $n$,  $X=x_1,\cdots, x_n$, the model $M$ is expected to produce a corresponding sequence of labels $\hat{Y}=\hat{y}_1, \cdots, \hat{y}_n$, where each label $\hat{y}_i \in L$ is associated with the linguistic feature of the token $x_i$.

In decomposed prompting, we design a prompt template function $T(\cdot, \cdot)$ which generates a specific prompt for each token. 
$T$ takes the input sequence $X$ and an individual token $x_i$ as arguments and returns a prompt for predicting the label of the token. The true label $y_i$ can be optionally included as an argument to $T$; if included, $T$ utilizes $y_i$ to provide a demonstration.

$C=c_1, \cdots, c_m$ is a sample from the training set. In the few-shot learning scenario, $k$ examples in the tuple format $(C_j, c_j, l_j)$ are given along with the input sequence $X$, where $c_j$ is a token in $C_j$, and $l_j \in L$ is the label for $c_j$. The demonstration $D$ of an input sequence $X$ is formulated as:
\begin{equation}
  \eqlabel{demon}
  D=I \circ\ T(C_1,c_1, l_1)\circ\ \cdots\circ\ T(C_k,c_k, l_k)
\end{equation} 
where $I$ denotes an optional instruction in natural language, $\circ$ denotes the string concatenation operation.
Finally, we use a prompt generator function $G(\cdot, \cdot)$ to create the set of decomposed prompts for an input sequence $X$:
\begin{equation}
  \eqlabel{gen}
  G(X, D)=\{D \circ\ T(X, x_1), \cdots, D \circ\ T(X, x_m)\}
\end{equation} 
The label $\hat{y_i}$ of token $x_i$ is predicted as follows:
\begin{equation}
  \eqlabel{pred}
  \hat{y_i}=\argmax\limits_{y\in L} P_M(l|D \circ\ T(X, x_i))
\end{equation}
For each possible label $y$, we obtain the probability that the model predicts this label as the next token and select the most likely label as the predicted label.

\section{Experiment and Results}
\subsection{Experimental Setup}

% In our study, we focus on evaluating the multilingual linguistic structure knowledge of English-centric models through multilingual part-of-speech tagging tasks, employing our proposed decomposed prompting method. 
\paragraph{Dataset and Language}
We use a subset of the Universal Dependency treebanks (UDPOS)~\citep{nivre-etal-2020-universal} to probe the multilingual linguistic knowledge of LLMs. The UDPOS dataset adopts a universal POS tag set consisting of 17 tags (\appref{app_tagset}). Our chosen subset, derived from the XTREME multilingual benchmark~\citep{hu2020xtreme}, comprises 38 languages from diverse language families distributions (\appref{langs}). We randomly sample 200 instances of each language for the evaluation.  

\paragraph{Model and Setup}
We experiment on three English-centric LLMs: LLaMA2-7B, LLaMA2-13B~\citep{touvron2023llama2}, and Mistral-7B~\citep{jiang2023mistral}, as well as two multilingual LLMs: BLOOMZ-7B~\citep{muennighoff-etal-2023-crosslingual} and mTk-Instruct~\citep{wang-etal-2022-super}.
All LLMs in our experiment are instruction-tuned versions accessible through the HuggingFace framework~\citep{wolf-etal-2020-transformers}.
We use the weighted average F1 scores for different tags as our evaluation metric. All experiments were conducted on a server with 4 \texttt{A100-SXM4-80GB} GPUs. More details of experimental settings are described in \appref{setup_details}.

% \paragraph{English-centric LLMs}
% \textbf{LLaMA2} represents an advanced iteration of the LLaMA foundation models developed by Meta AI~\citep{touvron2023llama, touvron2023llama2}, trained on publicly available corpora predominantly in English. 
% %Compared to its predecessor, LLaMA2 benefits from an enhanced data cleaning process, expanded language coverage, and the implementation of more efficient grouped-query attention~\citep{ainslie2023gqa}. 
% We consider LLaMA2 models with 7B and 13B parameters in our experiments.
% \textbf{Mistral 7B}~\citep{jiang2023mistral} enhances the LLaMA models in terms of both performance and inference efficiency, achieved through meticulous engineering in language model design and training.
% For our experiments, we utilize the instruction-tuned version of Mistral 7B, which has been fine-tuned on the OpenHermes 2.5 dataset~\footnote{\url{https://huggingface.co/datasets/teknium/OpenHermes-2.5}}. 

% \paragraph{Multilingual LLMs}
% \textbf{BLOOMZ}~\citep{muennighoff-etal-2023-crosslingual} is a multi-task fine-tuned variant of the BLOOM model~\citep{workshop2022bloom}, which is trained on 46 languages. We employ its 7B version in our experiment.
% \textbf{mTk-Instruct}~\citep{wang-etal-2022-super} is a multilingual encoder-decoder model, fine-tuned on instruction-following datasets. It is built upon the mT5 model~\citep{xue-etal-2021-mt5}, which is pretrained on corpora of over 100 languages. It comprises approximately 13 billion parameters.

% This dataset features instructions generated by GPT-4~\citep{achiam2023gpt, peng2023instruction}.

\subsection{Overall Results}
\seclabel{results}

\begin{table}[t]
\centering
\scalebox{0.7}{
\begin{tabular}{cc|cc|cc|c}
\toprule
          \multirow{2}{*}{\textbf{Model}}  & \multirow{2}{*}{\textbf{Method}}  & \multicolumn{2}{c|}{\textbf{Zero-shot}}   & \multicolumn{2}{c|}{\textbf{Few-shot}}  & \multirow{2}{*}{\textbf{Avg.}}     \\
\cline{3-6} 
            &  & \textbf{en}        & \textbf{mult.} & \textbf{en}       & \textbf{mult.} \\
\midrule
\multirow{2}{*}{\texttt{LLaMA2-7B}} &\textit{Iter}       & 33.1      & 27.2  & 68.0     & 48.6  & 44.2 \\
& \textit{Decom} & 58.2      & 43.2 & 74.7     & 50.5  & 56.7\\
\midrule
\multirow{2}{*}{\texttt{LLaMA2-13B}} &   \textit{Iter}   & 47.6      & 37.4  & 68.0     & 52.6  & 51.4\\
& \textit{Decom} & \textbf{67.3 }     & 54.7  & 77.3     & 54.5  & 63.5 \\
\midrule
\multirow{2}{*}{\texttt{Mistral-7B}}  & \textit{Iter}          & 65.2      & 54.3  & 80.2     & 58.9  & 64.7\\
 & \textit{Decom} & 63.6      & \textbf{61.8}  & \textbf{85.0}     & \textbf{64.4}  & \textbf{68.7}\\
 \midrule
\texttt{BLOOMZ-7B} & \textit{Decom} & 20.6      & 17.6 & 44.1     & 36.2  & 29.6 \\
\midrule
\texttt{mTK-Instruct} & \textit{Decom} & 47.6     & 43.1  & 57.3     & 44.7  & 48.2 \\
\bottomrule
\end{tabular}}
\caption{Overall results of iterative and decomposed prompting methods on POS tagging tasks in zero- and few-shot settings, with F1 score reported. \textbf{en} indicates the results for English, and \textbf{mult.} represents the average F1 score across other 37 languages. The best performance of each column is highlighted in \textbf{bold}.}
\tablabel{main_results}    
\end{table}

We evaluate the performance of \textit{iterative prompting}, the baseline method, and \textit{decomposed prompting}, our proposed method, for English and multilingual POS tag labeling tasks under zero- and few-shot settings. 
The few-shot examples and the prompts employed in our experiment are presented in \appref{prompt_details} for reference. Our preliminary experiment to explore the influence of the number of few-shot samples ($k$) reveals a mild impact on performance once $k$ increases to around 10. More details are provided in \appref{few_shot_study}.
% Our goal is (1) to validate the benefits of decomposed prompting in comparison to the baseline method (\secref{results}), and (2) to explore the extent to which decomposed prompting captures multilingual linguistic structure knowledge from the English-centric LLMs (\secref{analysis}).

\paragraph{Superiority in Efficacy}
The overall results for English-centric LLMs, as detailed in \tabref{main_results}, demonstrate that our proposed decomposed prompting obviously outperforms the iterative prompting baseline across both zero- and few-shot settings, in both English and multilingual evaluations. This trend holds true for all three English-centric models tested, with the sole exception in the zero-shot setting for the English evaluation with the Mistral-7B model, where \textit{Decom} slightly lags behind \textit{Iter} (63.6 vs. 65.2). In addition, English-centric LLMs outperform multilingual LLMs by a considerable margin. The complete experimental results are displayed in \appref{full_results}.

\begin{table}[h]
    \centering
    \scalebox{0.6}{
    \begin{tabular}{ccccc}
    \toprule
         & \textbf{BLOOMZ-7B} & \textbf{LLaMA2-7B} & \textbf{Mistral-7B} & \textbf{Avg.} \\
        \hline
        zero-shot & $3.2\times$ & $ 2.5\times$ & $ 1.4\times$ & $ 2.4\times$\\
        few-shot & $9.2\times $ & $7.9\times $ & $3.1\times $ & $6.7\times $\\
    \bottomrule
    \end{tabular}}
    \caption{The ratio by which the inference is accelerated for \textit{Decom} promoting compared to \textit{Iter} prompting. The inference speed was measured over the entire test set.}
    \tablabel{efficiency}
\end{table}

\paragraph{Superiority in Efficiency}
In addition to superior performance, \textit{decomposed prompting} offers enhanced efficiency during inference, especially in few-shot prompting. As demonstrated in \tabref{efficiency}, our proposed method achieves, on average, a 2.4-fold increase in speed compared to the baseline in the zero-shot prompting setting and a 6.7-fold increase in the few-shot setting. The efficiency advantage is less obvious with Mistral, owing to Mistral's implementation of a modified attention mechanism designed to enhance inference efficiency.

\section{Multilingual Analysis}
\seclabel{analysis}

\begin{figure}[t]
\includegraphics[width=\linewidth]{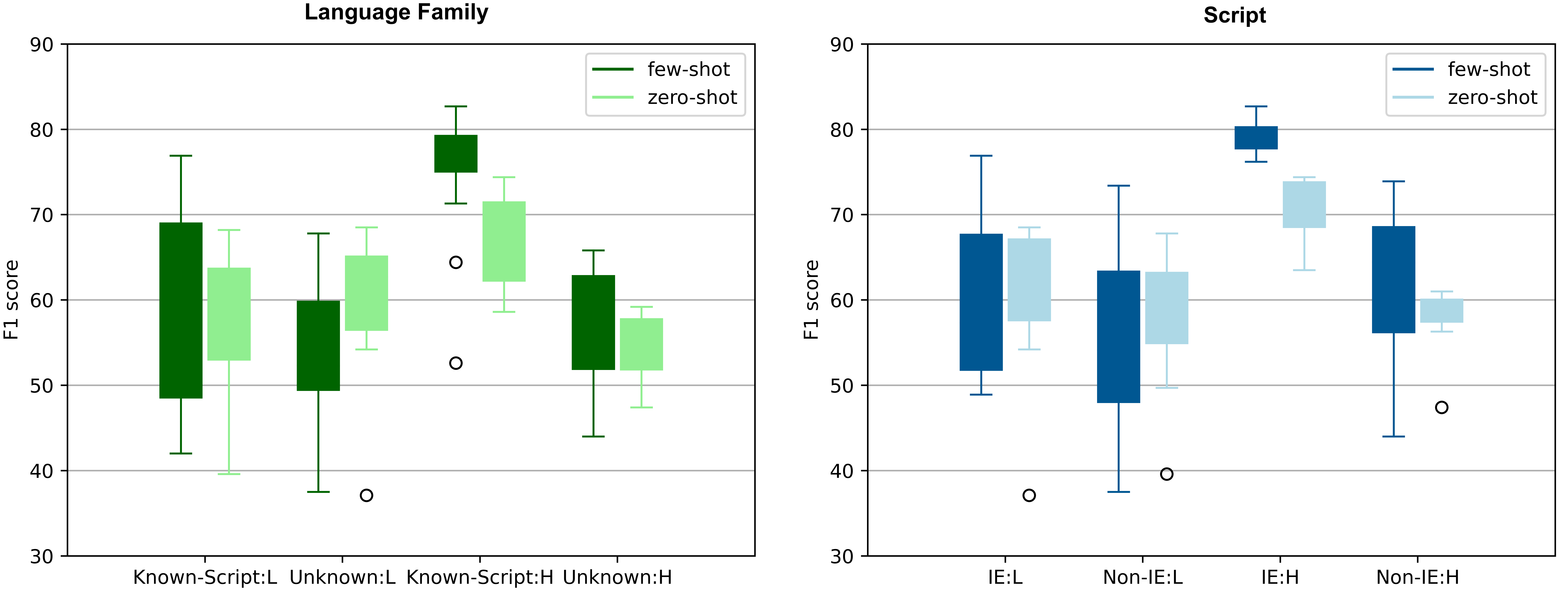}
	\caption{Analysis of decomposed promoting performance grouped by language family (a) and script type (b) under zero- and few-shot settings on Mistral. ``IE'' refers to the Indo-European language family. ``L'' (Low) represents languages that constitute less than 0.005\% of the pretraining corpus, while ``H'' (High) denotes all other languages.}
    \figlabel{box}
\end{figure}

\figref{box} provides a stratified view of decomposed prompting performance by language family and script, under both zero- and few-shot settings on the Mistral model. 
The results indicate that Indo-European languages generally achieve higher F1 scores compared to their non-Indo-European counterparts.
Notably, the presence of few-shot examples consistently improves the overall performance across all categories, but the box plot also shows that some languages are negatively impacted by the use of English demonstrations.
As discussed in \secref{background}, English-centric LLMs are adept at tokenizing words from Latin or Cyrillic scripts into subtokens. 
For scripts less familiar to these models, they often default to breaking down the text into UTF-8 encodings, which may lead to suboptimal representations for languages using these less common scripts. Thus, to capture a more nuanced understanding of LLM performance across linguistic varieties, we categorize languages not only by family but also by script type. \figref{box}(b) illustrates that, in both few-shot and zero-shot settings, languages with known scripts tend to yield better performance than unknown scripts. An exception to this trend is observed among the language group with smaller corpora in the zero-shot setting.

\begin{figure}[!t]
\includegraphics[width=\linewidth]{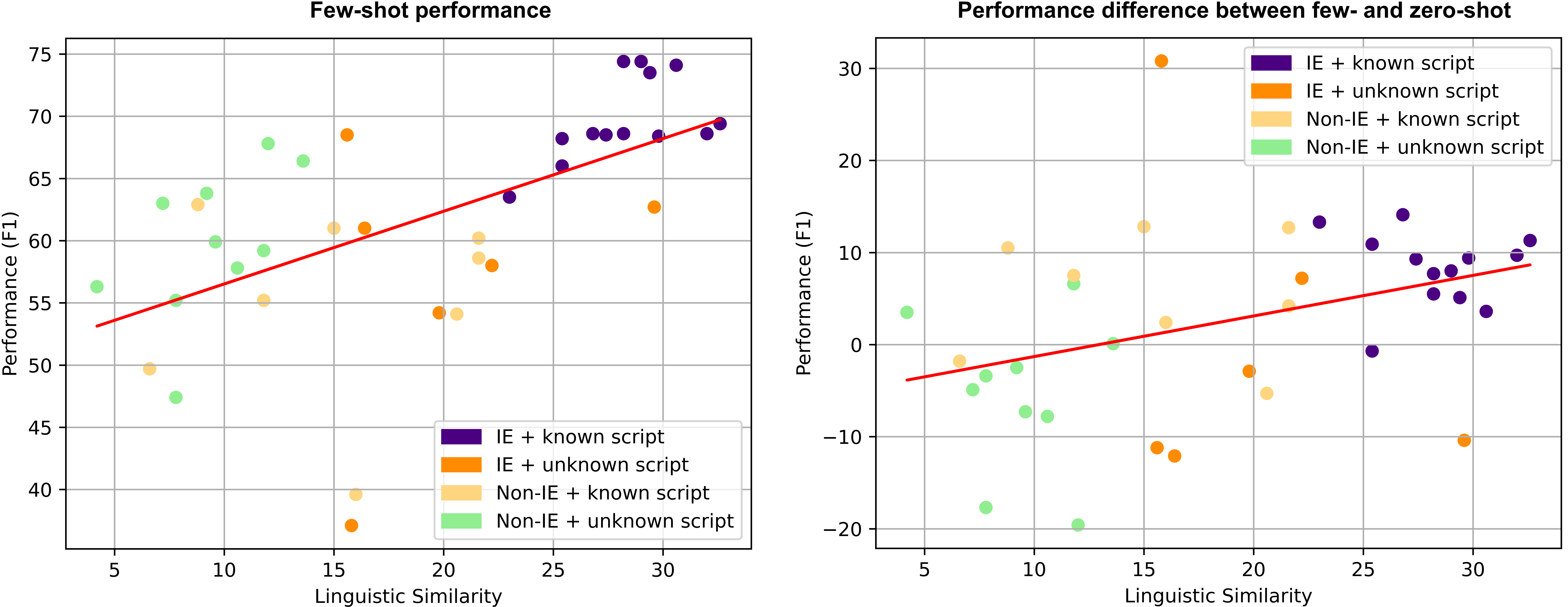}
	\caption{Panorama of Mistral model's per-language performance. Each node symbolizes a distinct language. (a) shows the few-shot performance and (b) shows the difference between few- and zero-shot performance for each language.}
    \figlabel{scatter}
\end{figure}

To further understand the impact of English demonstrations on languages with varied properties in multilingual prompting, we delve deeper into the cross-lingual transferability of English-centric LLMs and conduct a detailed analysis of individual language performance. We begin by quantifying the linguistic proximity of each tested language to English. This was achieved by calculating the cosine similarity between language vectors~\citep{littell-etal-2017-uriel} that incorporate syntactic, phylogenetic, and geographic attributes, among others, following~\citet{nie-etal-2023-cross} and \citet{ma2023promptbased}. Further information on the computation of language similarity is available in \appref{sim_computation}.
From \figref{scatter}, we observe that the performance gain from few-shot prompting is more substantial for languages that are linguistically closer to English, as indicated by the upward trend on the right side of the plot. Remarkably, languages distant from English may even experience a decline in performance when using English demonstrations.

\section{Conclusion}
In conclusion, we propose \textit{decomposed prompting}, a simple yet effective prompting method specially designed for sequence labeling tasks, addressing the difficulties of LLM benchmarking on sequence labeling tasks.
Our method outperforms iterative prompting techniques in terms of accuracy and efficiency in different experimental settings.
By applying \textit{decomposed prompting} to UDPOS dataset, we probe the multilingual linguistic structure knowledge of English-centric LLMs. 
Our multilingual investigation reveals that gain from few-shot decomposed prompting is generally more pronounced for languages closer to English.

\section*{Limitations}
Although our proposed \textit{decomposed prompting} method achieves overall remarkable performance in terms of both accuracy and efficiency, it has limitations for some special cases, for example, it can not well handle the case where the same word occurs twice in a sentence with different POS tags. Besides, the efficiency of decomposed prompting suffers as the length of the input sequence and the complexity of the task increase. Our study uses decomposed prompting methods for part-of-speech (POS) tagging as a means to evaluate the multilingual structural knowledge of English-centric Large Language Models (LLMs). This provides a foundational assessment of the models' capabilities. Nevertheless, extending the application scope of this methodology to probe more intricate aspects of linguistic structure is necessary. Future research could beneficially apply decomposed prompting to the analysis of complex linguistic phenomena, including sentence chunking and named entity recognition, to gain a deeper understanding of the nuanced capabilities of LLMs in processing and understanding language.

\section*{Acknowledgements}
We want to thank the anonymous reviewers for
their valuable feedback.
This work was supported
by the German Research Foundation (DFG) under grant SCHU 2246/14-1, Munich Center for Machine Learning (MCML), and China Scholarship Council (CSC).

\bibliography{anthology,latex/acl_latex}
% \bibliography{custom}

% \newpage

\appendix

\section{Experimental Setup Details}
Details of the experimental setup are introduced in this section.
\subsection{Dataset and Languages}

\subsubsection{POS Tag Set}
\applabel{app_tagset}
\figref{tagset} shows the pos tag set in UD. We also use the text in the box as the task instruction in our experiments.

\begin{figure}[ht]
    \centering
    \begin{tcolorbox}
    [colback=gray!20, colframe=gray!100, sharp corners, leftrule={2pt}, rightrule={0pt}, toprule={0pt}, bottomrule={0pt}, left={2pt}, right={2pt}, top={3pt}, bottom={2pt}]
{\small
\texttt{POS tag set: ADJ ADP ADV AUX CCONJ DET INTJ NOUN NUM PART PRON PROPN PUNCT SCONJ SYM VERB X}
}
\end{tcolorbox}
    \caption{UD POS tag set.}
    \figlabel{tagset}
\end{figure}

\subsubsection{Profile of Languages}
\applabel{langs}

As \figref{lang_pie} shows, our experiment involves 38 languages with diverse language family distributions.

\begin{figure}[ht]
    \centering
    \includegraphics[width=\linewidth]{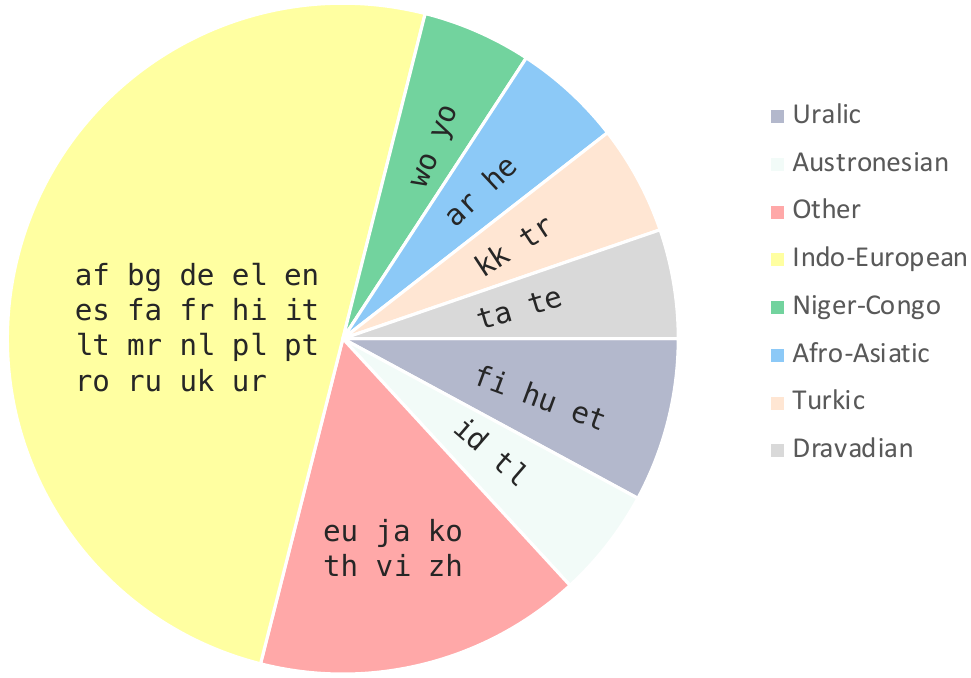}
    \caption{Distribution of languages by language family in the dataset.}
    \figlabel{lang_pie}
\end{figure}

\subsection{Baselines and Settings}
\applabel{setup_details}

\paragraph{Iterative Prompting (\textit{Iter})}
\citet{blevins-etal-2023-prompting} introduced a structured prompting approach that \textit{iteratively} labels an entire sentence by appending each predicted label to the context along with the subsequent word. This method is employed as a strong baseline in our study.

\paragraph{Decomposed Prompting (\textit{Decom})}
To evaluate our proposed approach, we employ the prompt template outlined in~\secref{method} to decompose the entire sequence into a set of individual prompts for prediction.
In our experiments, we use the 17 POS tags themselves as the label words, i.e., we expect the model to directly predict a tag from the tagset shown in \figref{tagset} by selecting the tag with the highest logit.

\paragraph{Zero- and Few-Shot Prompting}
We devised two experimental scenarios for multilingual prompting---zero-shot and few-shot---to evaluate the performance of both approaches under different conditions. In the zero-shot setting, only an English \textit{task instruction} is provided alongside the input in the target language. The text in~\figref{tagset}, which outlines the tag set information, serves as the instruction in our experiments. In few-shot prompting, we supplement the prompt with a few English demonstrations, structured according to the prompt template of each method. For \textit{Decom}, we randomly select an example for each tag type from the English training set to create a demonstration. For a fair comparison, the same number of demonstrations are used for the \textit{Iter} baseline.

% \paragraph{Evaluation Methods}
% We contrast two evaluation methodologies for prompting in our experiments. The \textit{probability-based} method leverages the model's output logits to retrieve the probability distribution over the tag set, subsequently identifying the label with the highest probability. In case the label word is tokenized into subtokens, we use the first subtoken to serve as the label word, following previous work \citep{pmlr-v139-zhao21c,wang-etal-2023-label}. The \textit{generation-based} method directly compares the content generated by the LLM with the gold label. 

\subsection{Language Similarity Computation}
\applabel{sim_computation}
\citet{malaviya-etal-2017-learning} and \citet{littell-etal-2017-uriel} proposed \textsc{lang2vec}, language vectors to represent various linguistic features for languages. A language can be represented by five vectors, containing syntactic, phonological, phonetic, phylogenetic, and geographical features, respectively. Linguistic similarities among different languages with respect to these linguistic features can be calculated through the cosine similarity.
In our study, we utilized the language vectors provided by \textsc{lang2vec} to calculate the cosine similarity between target languages and English. We used a rank-based similarity score to average the rank of languages in each feature dimension.
\tabref{lang_sim_full} illustrates the computation details.

\begin{table*}[b]
    \centering
    \scalebox{0.75}{
    \begin{tabular}{c|cc|cc|cc|cc|cc|c}
    \toprule
        & \textbf{syn.}  & \textbf{syn\_rank} & \textbf{pho. } & \textbf{pho\_rank} & \textbf{inv.}  & \textbf{inv\_rank} &\textbf{ fam.}  & \textbf{fam\_rank} & \textbf{geo.}  & \textbf{geo\_rank }& \textbf{rank\_score} \\
        \midrule
eng-nld & 92.43 & 37        & 81.83 & 18        & 76.28 & 36        & 44.51 & 35        & 99.96 & 37        & 32.6        \\
eng-deu & 90.26 & 36        & 80.60 & 15        & 78.68 & 37        & 54.49 & 37        & 99.76 & 35        & 32.0        \\
eng-ukr & 84.73 & 32        & 85.83 & 32        & 74.91 & 33        & 15.03 & 30        & 99.28 & 26        & 30.6        \\
eng-por & 84.24 & 31        & 90.46 & 35        & 74.03 & 28        & 10.14 & 22        & 99.68 & 33        & 29.8        \\
eng-ell & 78.31 & 25        & 95.35 & 37        & 74.74 & 32        & 15.03 & 32        & 98.96 & 22        & 29.6        \\
eng-pol & 78.64 & 26        & 85.83 & 29        & 74.09 & 29        & 15.03 & 31        & 99.63 & 32        & 29.4        \\
eng-bul & 85.78 & 35        & 85.83 & 30        & 74.38 & 30        & 13.73 & 27        & 99.01 & 23        & 29.0        \\
eng-ita & 85.78 & 34        & 85.83 & 28        & 72.94 & 26        & 11.21 & 23        & 99.53 & 30        & 28.2        \\
eng-rus & 81.18 & 29        & 85.83 & 31        & 74.63 & 31        & 16.80 & 33        & 95.81 & 17        & 28.2        \\
eng-ron & 79.60 & 27        & 90.46 & 34        & 73.42 & 27        & 11.89 & 24        & 99.22 & 25        & 27.4        \\
eng-spa & 82.16 & 30        & 85.83 & 27        & 72.83 & 25        & 9.71  & 21        & 99.59 & 31        & 26.8        \\
eng-lit & 69.33 & 18        & 80.42 & 14        & 75.58 & 34        & 19.39 & 34        & 99.44 & 27        & 25.4        \\
eng-afr & 84.94 & 33        & 81.83 & 17        & 75.91 & 35        & 50.46 & 36        & 86.84 & 6         & 25.4        \\
eng-fra & 81.18 & 28        & 75.28 & 7         & 72.24 & 24        & 9.71  & 20        & 99.93 & 36        & 23.0        \\
eng-est & 77.35 & 24        & 85.83 & 25        & 70.81 & 19        & 0.23  & 15        & 99.45 & 28        & 22.2        \\
eng-hun & 69.40 & 19        & 85.83 & 24        & 70.66 & 18        & 0.33  & 18        & 99.46 & 29        & 21.6        \\
eng-fin & 71.08 & 21        & 87.05 & 33        & 70.00 & 17        & 0.19  & 13        & 99.19 & 24        & 21.6        \\
eng-eus & 62.36 & 13        & 85.29 & 21        & 70.00 & 16        & 3.33  & 19        & 99.76 & 34        & 20.6        \\
eng-urd & 61.63 & 12        & 85.83 & 26        & 71.98 & 23        & 12.71 & 25        & 92.54 & 13        & 19.8        \\
eng-mar & 56.50 & 8         & 80.42 & 13        & 71.57 & 22        & 13.73 & 28        & 89.80 & 11        & 16.4        \\
eng-wol & 63.92 & 14        & 85.83 & 23        & 69.73 & 15        & 0.17  & 10        & 96.24 & 18        & 16.0        \\
eng-hin & 61.63 & 11        & 78.35 & 10        & 70.91 & 20        & 12.71 & 26        & 91.10 & 12        & 15.8        \\
eng-fas & 50.03 & 3         & 78.35 & 11        & 70.94 & 21        & 13.73 & 29        & 94.23 & 14        & 15.6        \\
eng-ind & 72.66 & 22        & 90.92 & 36        & 67.09 & 12        & 0.12  & 4         & 79.16 & 1         & 15.0        \\
eng-heb & 75.15 & 23        & 72.55 & 5         & 69.10 & 14        & 0.13  & 6         & 97.16 & 20        & 13.6        \\
eng-ara & 65.11 & 16        & 70.09 & 3         & 68.38 & 13        & 0.15  & 9         & 97.04 & 19        & 12.0        \\
eng-tur & 50.68 & 4         & 81.83 & 16        & 67.09 & 11        & 0.14  & 7         & 98.25 & 21        & 11.8        \\
eng-zho & 71.08 & 20        & 72.55 & 4         & 66.94 & 10        & 0.33  & 16        & 88.42 & 9         & 11.8        \\
eng-kaz & 44.77 & 1         & 83.64 & 19        & 66.59 & 9         & 0.14  & 8         & 95.22 & 16        & 10.6        \\
eng-vie & 66.04 & 17        & 78.35 & 9         & 65.81 & 8         & 0.19  & 11        & 85.25 & 3         & 9.6         \\
eng-tel & 52.07 & 6         & 80.42 & 12        & 64.76 & 4         & 0.19  & 14        & 89.18 & 10        & 9.2         \\
eng-tgl & 60.89 & 10        & 85.83 & 22        & 64.76 & 5         & 0.13  & 5         & 82.15 & 2         & 8.8         \\
eng-tam & 51.36 & 5         & 85.29 & 20        & 64.37 & 3         & 0.11  & 3         & 87.95 & 8         & 7.8         \\
eng-kor & 55.29 & 7         & 74.65 & 6         & 63.83 & 2         & 0.33  & 17        & 86.93 & 7         & 7.8         \\
eng-tha & 63.95 & 15        & 78.35 & 8         & 65.40 & 7         & 0.11  & 2         & 85.25 & 4         & 7.2         \\
eng-yor & 60.04 & 9         & 66.77 & 2         & 65.29 & 6         & 0.10  & 1         & 94.98 & 15        & 6.6         \\
eng-jpn & 50.03 & 2         & 66.77 & 1         & 56.88 & 1         & 0.19  & 12        & 85.65 & 5         & 4.2 \\
\bottomrule
    \end{tabular}}
    \caption{Details of language similarity computation.}
    \tablabel{lang_sim_full}
\end{table*}

\section{More Details of Decomposed Prompting Method}
\subsection{Intuition}
This method draws inspiration from the step-by-step thinking process humans employ when annotating linguistic features within a sentence. Typically, humans approach such tasks incrementally, addressing each token individually. Mirroring this intuitive strategy, our method first decomposes an input sentence into tokens. Subsequently, we generate a distinct prompt for each token, thereby transforming the sequence labeling task into a series of focused, manageable prompts.
\figref{example} illustrates the generation of sequence labeling prompts for the German sentence \textit{``Viel Erfolg!''} via \textit{decomposed prompting}.

\begin{figure}[ht]
    \centering
    \includegraphics[width=\linewidth]{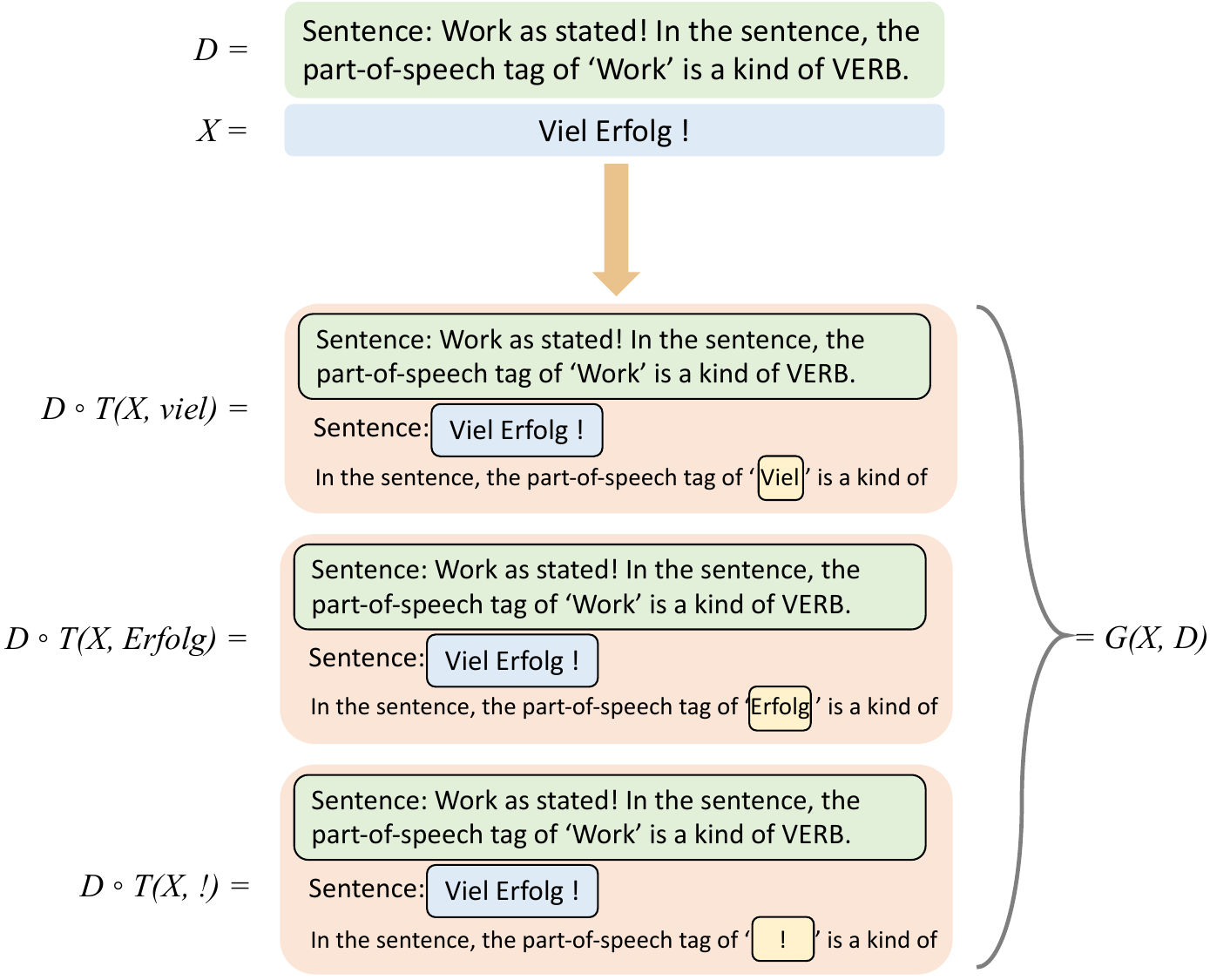}
    \caption{An example of how \textit{decomposed prompting} is implemented for sequence labeling.}
    \figlabel{example}
\end{figure}
\figref{example} illustrates the generation of sequence labeling prompts for the German sentence \textit{``Viel Erfolg!''} via \textit{decomposed prompting}.

An example of a template function is illustrated as follows.

\begin{tcolorbox}
    [colback=gray!20, colframe=gray!100, sharp corners, leftrule={3pt}, rightrule={0pt}, toprule={0pt}, bottomrule={0pt}, left={2pt}, right={2pt}, top={3pt}, bottom={3pt}]
{\small 
$T(X,x_i) = $ ``Sentence: $X$. In the sentence, the part-of-speech tag of \textquotesingle$ x_i$\textquotesingle\ is a kind of'' \\
$T(X,x_i, y_i) = $ ``Sentence: $X$. In the sentence, the part-of-speech tag of  \textquotesingle$x_i$\textquotesingle\ is a kind of $y_i$.''}
\end{tcolorbox}

\begin{figure}[!t]
    \centering
    \includegraphics[width=\linewidth]{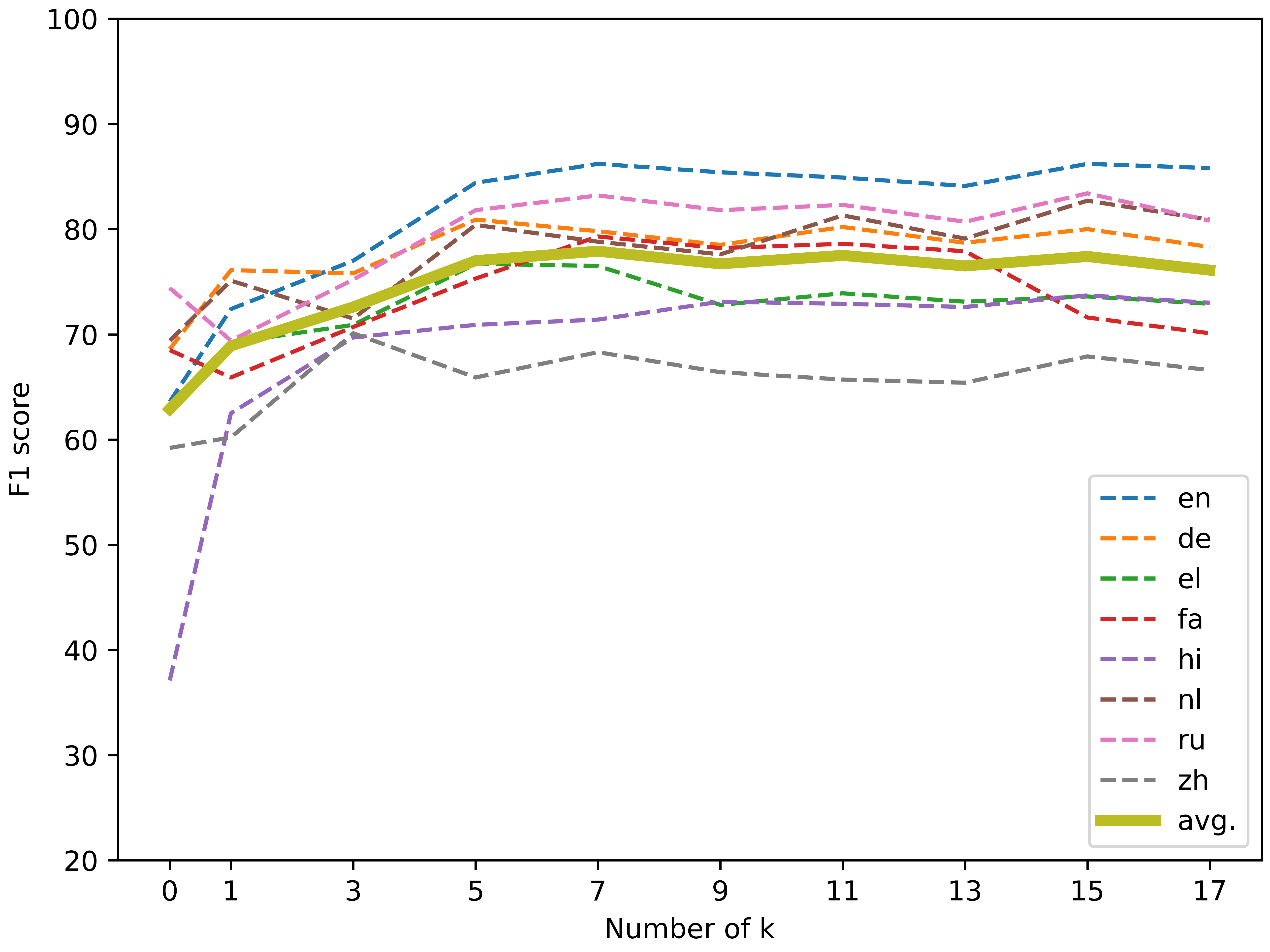}
    \caption{Performance dynamics with different numbers of few-shot samples. Experimental results of decomposed prompting with Mistral-7B.}
    \figlabel{few_shot}
\end{figure}

\subsection{Prompt Details}
\applabel{prompt_details}

Zero- and few-shot prompts used in this work are shown in \figref{prompt_decom} (decomposed prompting) and \figref{prompt_iter} (iterative prompting).
\begin{figure*}
    \centering
    \begin{tcolorbox}
    [colback=gray!20, colframe=gray!100, sharp corners, leftrule={2pt}, rightrule={0pt}, toprule={0pt}, bottomrule={0pt}, left={2pt}, right={2pt}, top={3pt}, bottom={2pt}]
{\footnotesize
\textbf{Zero-shot prompt}\\
\texttt{POS tag set: ADJ ADP ADV AUX CCONJ DET INTJ NOUN NUM PART PRON PROPN PUNCT SCONJ SYM VERB X \\
Sentence: Viel Erfolg ! \\
In the sentence, the part-of-speech tag of `Viel' is a kind of}\\
\textbf{Few-shot prompt} (w/o Instruction) \\
\texttt{Sentence: And if you send me a story , that would be great !\\
In the sentence, the part-of-speech tag of `if' is a kind of SCONJ.\\
Sentence: I `ll admit I was n't expecting much from this place , but they really did do a good job .\\
In the sentence, the part-of-speech tag of `good' is a kind of ADJ.\\
Sentence: I do n't know . The girl shrugged once again .
In the sentence, the part-of-speech tag of `girl' is a kind of NOUN.\\
Sentence: The dancers were falling back round a Polish agriculturalist who was teaching a gangling Englishman and two young Africans an Eastern European peasant dance .\\
In the sentence, the part-of-speech tag of `around' is a kind of ADP.\\
Sentence: Antigua was awesome .\\
In the sentence, the part-of-speech tag of `was' is a kind of AUX.\\
Sentence: The food is fresh and taste great .\\
In the sentence, the part-of-speech tag of `the' is a kind of DET.
Sentence: Now I have wife and son .\\
In the sentence, the part-of-speech tag of `Now' is a kind of ADV.\\
Sentence: However , this fruitful period was short-lived , as Greece suffered badly under the Ottoman Empire , only to recover in the 19th century as the capital of independent Greece .\\
In the sentence, the part-of-speech tag of `suffered' is a kind of VERB.\\
Sentence: I survived it without a problem .\\
In the sentence, the part-of-speech tag of `.' is a kind of PUNCT.
Sentence: The food is fresh and taste great .\\
In the sentence, the part-of-speech tag of `and' is a kind of CCONJ.\\
Sentence: you can view at dresscod.com\\
In the sentence, the part-of-speech tag of `dresscod.com' is a kind of X.\\
Sentence: I do n't know . The girl shrugged once again .\\
In the sentence, the part-of-speech tag of `I' is a kind of PRON.\\
Sentence: I `ll admit I was n't expecting much from this place , but they really did do a good job .\\
In the sentence, the part-of-speech tag of `n't' is a kind of PART.\\
Sentence: Antigua was awesome .\\
In the sentence, the part-of-speech tag of `Antigua' is a kind of PROPN.\\
Sentence: The dancers were falling back round a Polish agriculturalist who was teaching a gangling Englishman and two young Africans an Eastern European peasant dance .\\
In the sentence, the part-of-speech tag of `two' is a kind of NUM.
Sentence: Yes , the Cyclone is almost certain to lose strength as it surges over land .\\
In the sentence, the part-of-speech tag of `Yes' is a kind of INTJ.\\
Sentence: ----== Posted via Newsfeed.Com - Unlimited - Uncensored - Secure Usenet News ==----\\
In the sentence, the part-of-speech tag of `----== ` is a kind of SYM.\\
Sentence: Viel Erfolg !
In the sentence, the part-of-speech tag of `Viel' is a kind of}
}
\end{tcolorbox}
    \caption{Prompt design of decomposed prompting.}
    \figlabel{prompt_decom}
\end{figure*}
    
\begin{figure*}
    \centering
    \begin{tcolorbox}
    [colback=gray!20, colframe=gray!100, sharp corners, leftrule={2pt}, rightrule={0pt}, toprule={0pt}, bottomrule={0pt}, left={2pt}, right={2pt}, top={3pt}, bottom={2pt}]
{\footnotesize
\textbf{Zero-shot prompt}\\
\texttt{POS tag set: ADJ ADP ADV AUX CCONJ DET INTJ NOUN NUM PART PRON PROPN PUNCT SCONJ SYM VERB X \\
Sentence: Viel Erfolg ! \\
Viel\_}\\
\textbf{Few-shot prompt} (w/o Instruction) \\
\texttt{Context: Chahine said her immediate family spent about \$ 20,000 to return to Detroit via Syria and Jordan .\\
Tagged: Chahine\_PROPN said\_VERB her\_PRON immediate\_ADJ family\_NOUN spent\_VERB about\_ADV \$\_SYM 20,000\_NUM to\_PART return\_VERB to\_ADP Detroit\_PROPN via\_ADP Syria\_PROPN and\_CCONJ Jordan\_PROPN .\_PUNCT\\
Context: Welcome Darin !\\
Tagged: Welcome\_INTJ Darin\_PROPN !\_PUNCT\\
Context: you can view at dresscod.com\\
Tagged: you\_PRON can\_AUX view\_VERB at\_ADP dresscod.com\_X\\
$\cdots$ \\
Context: They work on Wall Street , after all , so when they hear a company who's stated goals include " Do n't be evil , " they imagine a company who's eventually history will be " Do n't be profitable . "\\
Tagged: They\_PRON work\_VERB on\_ADP Wall\_PROPN Street\_PROPN ,\_PUNCT after\_ADV all\_ADV ,\_PUNCT so\_ADV when\_ADV they\_PRON hear\_VERB a\_DET company\_NOUN who's\_PRON stated\_VERB goals\_NOUN include\_VERB "\_PUNCT Do\_AUX n't\_PART be\_AUX evil\_ADJ ,\_PUNCT "\_PUNCT they\_PRON imagine\_VERB a\_DET company\_NOUN who's\_PRON eventually\_ADJ history\_NOUN will\_AUX be\_VERB "\_PUNCT Do\_AUX n't\_PART be\_AUX profitable\_ADJ .\_PUNCT "\_PUNCT\\
Context: It 's not quite as freewheeling an environment as you 'd imagine : Sergey Brin has actually created a mathematical ' proof ' that the company 's self - driven research strategy , which gives employees one day a week to do research projects on their own , is a good , respectable idea .\\
Tagged: It\_PRON 's\_AUX not\_PART quite\_ADV as\_ADV freewheeling\_ADJ an\_DET environment\_NOUN as\_SCONJ you\_PRON 'd\_AUX imagine\_VERB :\_PUNCT Sergey\_PROPN Brin\_PROPN has\_AUX actually\_ADV created\_VERB a\_DET mathematical\_ADJ '\_PUNCT proof\_NOUN '\_PUNCT that\_SCONJ the\_DET company\_NOUN 's\_PART self\_NOUN -\_PUNCT driven\_VERB research\_NOUN strategy\_NOUN ,\_PUNCT which\_PRON gives\_VERB employees\_NOUN one\_NUM day\_NOUN a\_DET week\_NOUN to\_PART do\_VERB research\_NOUN projects\_NOUN on\_ADP their\_PRON own\_ADJ ,\_PUNCT is\_AUX a\_DET good\_ADJ ,\_PUNCT respectable\_ADJ idea\_NOUN .\_PUNCT\\
Context: Read the entire article ; there 's a punchline , too .\\
Tagged: Read\_VERB the\_DET entire\_ADJ article\_NOUN ;\_PUNCT there\_PRON 's\_VERB a\_DET punchline\_NOUN ,\_PUNCT too\_ADV .\_PUNCT\\
Context: My opinion piece on the implications of Arafat 's passing for al - Qaeda has appeared at Newsday .\\
Tagged: My\_PRON opinion\_NOUN piece\_NOUN on\_ADP the\_DET implications\_NOUN of\_ADP Arafat\_PROPN 's\_PART passing\_NOUN for\_ADP al\_PROPN -\_PUNCT Qaeda\_PROPN has\_AUX appeared\_VERB at\_ADP Newsday\_PROPN .\_PUNCT\\
Context: Viel Erfolg !
Tagged: Viel\_
}
}
\end{tcolorbox}
    \caption{Prompt design of iterative prompting.}
    \figlabel{prompt_iter}
\end{figure*}

\section{Few-Shot Ablation Study}
\applabel{few_shot_study}
we investigate the impact of the number of few-shot examples on the performance in the decomposed prompting. We randomly select 8 languages (en, de, el, fa, hi, hl, ru, zh) and explore their performance dynamics with the increasing of the few-shot samples. \figref{few_shot} shows that overall, when $k$ is small, increasing the number of samples bring performance improvement. As $k$ continues to increase, the performance tends to be stable and even gets worse when samples are too many. 

\section{Full Results}
\applabel{full_results}

Full experimental results are displayed in \tabref{Mistral-7b_results} (Mistral 7B), \tabref{Llama2-7b} (LLaMA2 7B), \tabref{Llama2-13b_result} (LLaMA 13B), \tabref{Bloomz-7b1_results} (BLOOMZ 7B), \tabref{mtk-11b_results} (mTk 13B), and \tabref{few_shot_tab} (few-shot ablation study).

\begin{table*}[ht]
\centering
\scalebox{0.75}{
\begin{tabular}{lllllllllllllll} 
\toprule
\multicolumn{2}{c}{language}                   & en   & af   & ar   & bg   & de   & el   & es   & et   & eu   & fa   & fi   & fr   & he    \\ 
\midrule
\multirow{3}{*}{zero-shot} & Iter              & 65.2 & 67.8 & 57.2 & 68.6 & 65.0 & 55.0 & 64.8 & 49.4 & 35.6 & 58.3 & 50.2 & 65.4 & 51.5  \\
                           & Decom (prob.)     & 63.6 & 66.0 & 67.8 & 74.4 & 68.6 & 62.7 & 68.6 & 58.0 & 54.1 & 68.5 & 60.2 & 63.5 & 66.4  \\
                           & Decom (gen.)      & 45.3 & 43.8 & 49.6 & 50.5 & 49.0 & 50.7 & 43.3 & 53.6 & 50.7 & 56.0 & 55.5 & 40.5 & 55.6  \\ 
\midrule
\multirow{5}{*}{few-shot}  & Iter              & 80.2 & 66.4 & 65.0 & 77.3 & 66.9 & 56.4 & 70.8 & 53.7 & 50.7 & 57.4 & 63.9 & 67.7 & 66.4  \\
                           & Decom (prob.)     & 85.0 & 76.9 & 48.1 & 82.4 & 78.3 & 52.3 & 82.7 & 65.2 & 48.8 & 57.3 & 64.4 & 76.9 & 66.6  \\
                           & Decom (gen.)      & 81.4 & 74.8 & 44.3 & 80.4 & 77.0 & 46.3 & 82.0 & 64.0 & 48.1 & 54.1 & 63.6 & 76.4 & 64.9  \\
                           & Decom (prob.) + I & 83.4 & 77.9 & 42.4 & 76.9 & 77.8 & 33.6 & 77.6 & 64.6 & 57.4 & 42.9 & 67.6 & 74.8 & 58.5  \\
                           & Decom (gen.) + I  & 78.7 & 75.8 & 34.0 & 74.9 & 76.6 & 24.7 & 76.4 & 62.6 & 56.8 & 34.4 & 64.5 & 73.4 & 54.5  \\ 
\midrule
\multicolumn{2}{c}{language}                   & hi   & hu   & id   & it   & ja   & kk   & ko   & lt   & mr   & nl   & pl   & pt   & ro    \\ 
\midrule
\multirow{3}{*}{zero-shot} & Iter              & 61.3 & 50.6 & 54.7 & 64.0 & 42.2 & 36.7 & 39.9 & 52.8 & 39.1 & 60.4 & 66.5 & 63.9 & 66.2  \\
                           & Decom (prob.)     & 37.1 & 58.6 & 61.0 & 68.6 & 56.3 & 57.8 & 47.4 & 68.2 & 61.0 & 69.4 & 73.5 & 68.4 & 68.5  \\
                           & Decom (gen.)      & 35.6 & 46.7 & 41.8 & 45.1 & 48.9 & 50.2 & 42.2 & 60.3 & 56.7 & 46.8 & 59.5 & 43.1 & 44.6  \\ 
\midrule
\multirow{5}{*}{few-shot}  & Iter              & 65.7 & 50.4 & 70.0 & 67.2 & 42.0 & 43.8 & 42.6 & 63.2 & 54.4 & 66.6 & 70.9 & 75.1 & 65.9  \\
                           & Decom (prob.)     & 67.8 & 71.3 & 73.9 & 76.2 & 59.8 & 50.0 & 44.0 & 67.5 & 48.9 & 80.6 & 78.6 & 77.8 & 77.8  \\
                           & Decom (gen.)      & 66.2 & 70.8 & 73.0 & 76.0 & 57.1 & 50.2 & 43.4 & 67.1 & 48.9 & 77.2 & 78.3 & 76.9 & 77.0  \\
                           & Decom (prob.) + I & 57.6 & 66.5 & 70.4 & 72.2 & 54.2 & 58.4 & 49.2 & 69.9 & 53.1 & 78.5 & 76.7 & 75.0 & 76.4  \\
                           & Decom (gen.) + I  & 55.3 & 63.9 & 68.2 & 70.3 & 53.1 & 57.9 & 48.2 & 69.5 & 52.7 & 76.9 & 75.7 & 74.2 & 75.1  \\ 
\midrule
\multicolumn{2}{c}{language}                   & ru   & ta   & te   & th   & tl   & tr   & uk   & ur   & vi   & wo   & yo   & zh   & avg.  \\ 
\midrule
\multirow{3}{*}{zero-shot} & Iter              & 68.2 & 39.2 & 51.1 & 54.1 & 65.0 & 47.7 & 67.0 & 56.0 & 41.7 & 31.5 & 41.3 & 58.8 & 54.3  \\
                           & Decom (prob.)     & 74.4 & 55.2 & 63.8 & 63.0 & 62.9 & 55.2 & 74.1 & 54.2 & 59.9 & 39.6 & 49.7 & 59.2 & 61.8  \\
                           & Decom (gen.)      & 54.7 & 52.2 & 57.4 & 50.1 & 51.3 & 43.2 & 57.4 & 40.3 & 45.9 & 29.2 & 43.3 & 55.7 & 48.7  \\ 
\midrule
\multirow{5}{*}{few-shot}  & Iter              & 74.0 & 52.0 & 62.4 & 57.1 & 37.3 & 62.0 & 68.2 & 59.6 & 41.0 & 25.2 & 39.0 & 62.3 & 58.9  \\
                           & Decom (prob.)     & 79.9 & 37.5 & 61.4 & 58.2 & 73.4 & 62.7 & 77.7 & 51.3 & 52.6 & 42.0 & 47.8 & 65.8 & 64.4  \\
                           & Decom (gen.)      & 78.0 & 33.9 & 61.3 & 56.9 & 73.4 & 62.6 & 76.2 & 45.7 & 52.8 & 42.0 & 47.6 & 64.5 & 63.0  \\
                           & Decom (prob.) + I & 76.8 & 35.7 & 67.0 & 45.8 & 74.9 & 63.7 & 75.1 & 40.5 & 59.4 & 43.1 & 49.2 & 62.9 & 62.3  \\
                           & Decom (gen.) + I  & 73.9 & 28.0 & 66.6 & 42.9 & 74.9 & 62.6 & 73.4 & 32.9 & 59.7 & 43.2 & 48.6 & 61.4 & 59.9  \\
\bottomrule
\end{tabular}}
\caption{Full results on Mistral 7b.}
\tablabel{Mistral-7b_results}
\end{table*}

\begin{table*}[ht]
\centering
\scalebox{0.75}{
\begin{tabular}{lllllllllllllll} 
\toprule
\multicolumn{2}{c}{language}                   & en   & af   & ar   & bg   & de   & el   & es   & et   & eu   & fa   & fi   & fr   & he    \\ 
\midrule
\multirow{3}{*}{zero-shot} & Iter              & 33.1 & 38.8 & 30.2 & 33.2 & 34.5 & 38.1 & 38.9 & 19.7 & 11.8 & 17.7 & 26.0 & 37.5 & 21.3  \\
                           & Decom (prob.)     & 58.2 & 45.1 & 49.6 & 55.9 & 53.3 & 50.4 & 44.7 & 37.7 & 36.4 & 40.5 & 41.3 & 46.8 & 39.5  \\
                           & Decom (gen.)      & 53.8 & 46.8 & 38.5 & 45.8 & 57.1 & 54.3 & 52.4 & 28.6 & 20.2 & 35.9 & 39.8 & 53.1 & 37.5  \\ 
\midrule
\multirow{5}{*}{few-shot}  & Iter              & 68.0 & 56.1 & 58.0 & 63.4 & 56.9 & 48.7 & 55.3 & 46.5 & 41.3 & 51.1 & 50.5 & 54.2 & 54.0  \\
                           & Decom (prob.)     & 74.7 & 60.0 & 29.9 & 64.7 & 63.0 & 30.6 & 55.7 & 53.0 & 44.4 & 29.7 & 62.9 & 54.4 & 42.8  \\
                           & Decom (gen.)      & 62.1 & 51.0 & 25.7 & 60.3 & 52.4 & 23.9 & 50.3 & 48.3 & 42.9 & 26.0 & 56.8 & 49.5 & 37.5  \\
                           & Decom (prob.) + I & 68.2 & 55.9 & 23.7 & 61.6 & 61.0 & 20.2 & 52.5 & 43.2 & 40.8 & 22.7 & 49.4 & 54.8 & 35.4  \\
                           & Decom (gen.) + I  & 63.4 & 53.2 & 19.0 & 57.9 & 56.2 & 12.0 & 47.8 & 39.3 & 40.0 & 15.5 & 46.4 & 51.2 & 30.1  \\ 
\midrule
\multicolumn{2}{c}{language}                   & hi   & hu   & id   & it   & ja   & kk   & ko   & lt   & mr   & nl   & pl   & pt   & ro    \\ 
\midrule
\multirow{3}{*}{zero-shot} & Iter              & 35.2 & 29.3 & 31.1 & 35.1 & 28.7 & 13.6 & 19.8 & 24.9 & 13.2 & 37.5 & 37.7 & 38.4 & 32.0  \\
                           & Decom (prob.)     & 36.9 & 47.0 & 46.9 & 46.7 & 32.4 & 39.0 & 29.0 & 34.9 & 45.3 & 54.9 & 54.0 & 48.6 & 43.6  \\
                           & Decom (gen.)      & 34.8 & 47.4 & 39.1 & 45.2 & 30.9 & 33.0 & 33.2 & 37.7 & 42.0 & 51.1 & 44.1 & 48.5 & 42.6  \\ 
\midrule
\multirow{5}{*}{few-shot}  & Iter              & 54.0 & 41.0 & 51.3 & 49.6 & 40.0 & 43.2 & 25.0 & 52.5 & 50.3 & 52.2 & 52.4 & 52.0 & 53.8  \\
                           & Decom (prob.)     & 45.8 & 62.6 & 60.9 & 56.4 & 40.2 & 51.4 & 48.2 & 56.3 & 47.3 & 58.9 & 67.2 & 60.3 & 63.6  \\
                           & Decom (gen.)      & 42.4 & 57.0 & 56.5 & 51.6 & 34.1 & 47.5 & 44.7 & 51.7 & 43.5 & 51.3 & 64.2 & 54.5 & 55.5  \\
                           & Decom (prob.) + I & 30.6 & 52.3 & 54.1 & 51.3 & 37.3 & 46.6 & 41.9 & 46.5 & 45.7 & 64.2 & 65.4 & 55.2 & 56.4  \\
                           & Decom (gen.) + I  & 24.1 & 50.6 & 49.5 & 44.1 & 32.9 & 46.0 & 40.7 & 45.3 & 34.5 & 60.2 & 62.0 & 51.2 & 51.8  \\ 
\midrule
\multicolumn{2}{c}{language}                   & ru   & ta   & te   & th   & tl   & tr   & uk   & ur   & vi   & wo   & yo   & zh   & avg.  \\ 
\midrule
\multirow{3}{*}{zero-shot} & Iter              & 29.8 & 19.2 & 13.8 & 29.2 & 28.6 & 22.2 & 30.3 & 20.7 & 29.7 & 13.3 & 13.7 & 32.2 & 27.2  \\
                           & Decom (prob.)     & 55.8 & 38.0 & 34.0 & 37.5 & 57.3 & 48.3 & 57.4 & 31.6 & 39.5 & 27.6 & 29.1 & 42.9 & 43.2  \\
                           & Decom (gen.)      & 48.7 & 25.5 & 36.9 & 34.6 & 66.3 & 45.9 & 48.8 & 28.4 & 35.3 & 18.7 & 21.8 & 44.0 & 40.4  \\ 
\midrule
\multirow{5}{*}{few-shot}  & Iter              & 58.2 & 30.9 & 54.3 & 49.4 & 37.3 & 34.4 & 57.7 & 44.0 & 46.5 & 40.7 & 39.3 & 52.0 & 48.6  \\
                           & Decom (prob.)     & 67.2 & 31.7 & 44.7 & 36.5 & 46.8 & 58.1 & 62.9 & 27.1 & 41.4 & 39.9 & 37.1 & 64.8 & 50.5  \\
                           & Decom (gen.)      & 62.3 & 25.3 & 43.5 & 34.7 & 45.4 & 55.9 & 59.4 & 23.7 & 40.7 & 36.2 & 35.5 & 50.9 & 45.8  \\
                           & Decom (prob.) + I & 59.6 & 20.3 & 38.4 & 20.9 & 63.1 & 54.1 & 59.9 & 19.3 & 49.7 & 32.2 & 33.8 & 48.2 & 45.1  \\
                           & Decom (gen.) + I  & 56.9 & 12.5 & 34.5 & 16.7 & 58.8 & 52.7 & 57.5 & 13.0 & 47.8 & 29.7 & 31.7 & 44.2 & 41.0  \\
\bottomrule
\end{tabular}}
\caption{Full results on LLaMA2 7b.}
\tablabel{Llama2-7b}
\end{table*}

\begin{table*}
\centering
\scalebox{0.75}{
\begin{tabular}{lllllllllllllll} 
\toprule
\multicolumn{2}{c}{language}                   & en   & af   & ar   & bg   & de   & el   & es   & et   & eu   & fa   & fi   & fr   & he    \\ 
\midrule
\multirow{3}{*}{zero-shot} & Iter              & 47.6 & 37.4 & 43.2 & 44.5 & 45.7 & 38.4 & 46.8 & 37.0 & 26.5 & 42.0 & 40.7 & 45.5 & 40.0  \\
                           & Decom (prob.)     & 67.3 & 60.1 & 54.4 & 62.7 & 63.6 & 60.5 & 55.9 & 49.9 & 37.4 & 59.8 & 62.6 & 53.4 & 55.4  \\
                           & Decom (gen.)      & 59.2 & 54.1 & 45.0 & 52.5 & 57.5 & 51.3 & 56.3 & 37.6 & 36.7 & 49.7 & 50.2 & 54.7 & 44.3  \\ 
\midrule
\multirow{5}{*}{few-shot}  & Iter              & 68.0 & 62.3 & 57.4 & 69.9 & 60.3 & 57.9 & 66.7 & 44.8 & 41.0 & 49.1 & 54.2 & 63.2 & 59.8  \\
                           & Decom (prob.)     & 77.3 & 67.8 & 33.2 & 67.6 & 67.5 & 35.0 & 62.6 & 58.5 & 46.9 & 34.7 & 62.8 & 64.8 & 48.4  \\
                           & Decom (gen.)      & 65.3 & 59.1 & 25.1 & 61.3 & 58.6 & 24.6 & 53.5 & 51.8 & 45.8 & 27.4 & 55.4 & 55.9 & 43.9  \\
                           & Decom (prob.) + I & 74.3 & 67.6 & 25.9 & 60.7 & 70.5 & 21.5 & 59.1 & 51.4 & 44.1 & 21.8 & 59.1 & 63.1 & 40.3  \\
                           & Decom (gen.) + I  & 68.7 & 64.4 & 19.2 & 58.7 & 66.2 & 12.4 & 53.9 & 47.9 & 42.2 & 15.5 & 54.0 & 59.7 & 35.0  \\ 
\midrule
\multicolumn{2}{c}{language}                   & hi   & hu   & id   & it   & ja   & kk   & ko   & lt   & mr   & nl   & pl   & pt   & ro    \\ 
\midrule
\multirow{3}{*}{zero-shot} & Iter              & 45.0 & 38.8 & 40.9 & 41.8 & 42.8 & 24.1 & 29.8 & 41.2 & 30.5 & 36.6 & 42.2 & 43.3 & 43.1  \\
                           & Decom (prob.)     & 53.8 & 57.6 & 57.4 & 54.8 & 48.3 & 51.8 & 45.1 & 54.3 & 50.2 & 62.0 & 66.4 & 56.6 & 57.9  \\
                           & Decom (gen.)      & 45.4 & 47.9 & 48.2 & 51.3 & 35.9 & 48.7 & 35.3 & 43.2 & 48.7 & 56.9 & 58.2 & 51.3 & 51.4  \\ 
\midrule
\multirow{5}{*}{few-shot}  & Iter              & 51.6 & 46.1 & 60.8 & 62.7 & 46.5 & 32.0 & 26.6 & 50.8 & 52.7 & 61.0 & 64.4 & 68.9 & 58.9  \\
                           & Decom (prob.)     & 45.4 & 69.8 & 62.2 & 61.2 & 44.6 & 52.3 & 46.1 & 63.0 & 49.6 & 65.4 & 68.1 & 62.3 & 63.6  \\
                           & Decom (gen.)      & 37.3 & 60.5 & 55.8 & 54.5 & 40.7 & 49.4 & 42.6 & 58.4 & 46.9 & 54.9 & 61.4 & 54.3 & 54.9  \\
                           & Decom (prob.) + I & 31.4 & 64.2 & 55.3 & 55.3 & 38.1 & 51.7 & 47.1 & 58.9 & 52.5 & 65.4 & 60.2 & 56.3 & 60.4  \\
                           & Decom (gen.) + I  & 23.4 & 60.0 & 50.2 & 52.4 & 35.5 & 49.0 & 45.3 & 56.9 & 50.8 & 61.1 & 58.2 & 54.1 & 56.1  \\ 
\midrule
\multicolumn{2}{c}{language}                   & ru   & ta   & te   & th   & tl   & tr   & uk   & ur   & vi   & wo   & yo   & zh   & avg.  \\ 
\midrule
\multirow{3}{*}{zero-shot} & Iter              & 42.6 & 21.8 & 22.5 & 45.6 & 29.3 & 29.9 & 39.8 & 35.1 & 36.0 & 24.4 & 24.1 & 45.2 & 37.4  \\
                           & Decom (prob.)     & 66.5 & 49.1 & 50.8 & 44.6 & 66.5 & 56.9 & 65.7 & 47.2 & 45.3 & 34.5 & 47.7 & 58.7 & 54.7  \\
                           & Decom (gen.)      & 55.2 & 46.2 & 54.1 & 44.2 & 73.1 & 52.8 & 57.3 & 40.2 & 45.4 & 29.9 & 39.6 & 52.5 & 48.7  \\ 
\midrule
\multirow{5}{*}{few-shot}  & Iter              & 64.9 & 33.5 & 51.5 & 51.5 & 60.2 & 46.3 & 61.6 & 45.4 & 41.8 & 36.3 & 31.6 & 52.1 & 52.6  \\
                           & Decom (prob.)     & 71.0 & 30.4 & 54.4 & 40.1 & 74.0 & 54.1 & 69.0 & 30.1 & 47.5 & 39.4 & 36.2 & 66.6 & 54.5  \\
                           & Decom (gen.)      & 63.3 & 21.9 & 51.3 & 33.9 & 70.9 & 52.2 & 61.4 & 22.1 & 45.2 & 38.1 & 34.8 & 56.5 & 48.3  \\
                           & Decom (prob.) + I & 63.3 & 22.3 & 52.2 & 23.5 & 70.7 & 53.9 & 62.4 & 19.0 & 48.4 & 36.9 & 36.4 & 56.7 & 49.4  \\
                           & Decom (gen.) + I  & 59.8 & 14.1 & 48.4 & 18.5 & 70.2 & 53.2 & 59.1 & 12.0 & 47.1 & 34.5 & 34.5 & 52.7 & 45.6  \\
\bottomrule
\end{tabular}}
\caption{Full results on LLaMA2 13b.}
\tablabel{Llama2-13b_result}
\end{table*}

\begin{table*}
\centering
\scalebox{0.75}{
\begin{tabular}{lllllllllllllll} 
\toprule
\multicolumn{2}{c}{language}                   & en   & af   & ar   & bg   & de   & el   & es   & et   & eu   & fa   & fi   & fr   & he    \\ 
\midrule
\multirow{3}{*}{zero-shot} & Iter              & 6.4  & 7.2  & 10.9 & 7.6  & 9.5  & 8.4  & 8.2  & 12.4 & 7.5  & 7.3  & 9.3  & 9.0  & 9.6   \\
                           & Decom (prob.)     & 20.6 & 20.5 & 14.5 & 19.7 & 26.2 & 18.3 & 18.2 & 22.3 & 19.0 & 12.8 & 19.2 & 19.4 & 15.2  \\
                           & Decom (gen.)      & 28.7 & 18.3 & 16.4 & 22.6 & 26.8 & 22.7 & 24.9 & 21.2 & 25.0 & 11.3 & 20.9 & 20.9 & 21.8  \\ 
\midrule
\multirow{5}{*}{few-shot}  & Iter              & 30.9 & 6.4  & 14.4 & 23.8 & 19.3 & 7.7  & 23.2 & 16.6 & 28.4 & 11.1 & 22.3 & 25.1 & 7.5   \\
                           & Decom (prob.)     & 44.1 & 33.1 & 28.7 & 35.9 & 44.0 & 39.2 & 33.6 & 39.0 & 38.4 & 25.6 & 38.5 & 35.6 & 34.3  \\
                           & Decom (gen.)      & 40.6 & 31.0 & 25.5 & 31.4 & 39.5 & 35.8 & 30.5 & 36.9 & 33.8 & 21.6 & 36.8 & 31.0 & 33.6  \\
                           & Decom (prob.) + I & 33.3 & 24.7 & 27.2 & 35.2 & 30.0 & 31.0 & 30.1 & 36.5 & 37.4 & 24.7 & 34.4 & 29.0 & 29.2  \\
                           & Decom (gen.) + I  & 33.3 & 24.5 & 27.1 & 35.0 & 29.7 & 30.4 & 30.0 & 36.4 & 37.1 & 24.5 & 34.5 & 28.9 & 29.1  \\ 
\midrule
\multicolumn{2}{c}{language}                   & hi   & hu   & id   & it   & ja   & kk   & ko   & lt   & mr   & nl   & pl   & pt   & ro    \\ 
\midrule
\multirow{3}{*}{zero-shot} & Iter              & 3.9  & 13.0 & 10.0 & 9.1  & 2.8  & 4.5  & 8.5  & 7.8  & 0.4  & 9.1  & 9.9  & 8.6  & 8.8   \\
                           & Decom (prob.)     & 12.0 & 27.0 & 17.7 & 23.1 & 13.5 & 17.7 & 19.5 & 23.6 & 12.4 & 18.6 & 23.6 & 19.5 & 19.6  \\
                           & Decom (gen.)      & 15.2 & 21.9 & 17.3 & 26.2 & 26.2 & 16.8 & 21.3 & 23.4 & 25.8 & 14.7 & 23.2 & 27.8 & 24.3  \\ 
\midrule
\multirow{5}{*}{few-shot}  & Iter              & 20.5 & 13.4 & 30.5 & 19.0 & 6.3  & 17.0 & 5.9  & 15.0 & 35.2 & 20.8 & 17.9 & 27.4 & 13.4  \\
                           & Decom (prob.)     & 27.0 & 38.2 & 43.8 & 33.9 & 25.9 & 45.6 & 35.0 & 40.3 & 39.6 & 39.8 & 39.7 & 34.4 & 33.3  \\
                           & Decom (gen.)      & 24.8 & 36.9 & 41.2 & 31.1 & 22.5 & 43.8 & 32.7 & 39.5 & 28.0 & 36.5 & 36.5 & 31.7 & 32.0  \\
                           & Decom (prob.) + I & 25.6 & 32.3 & 36.0 & 30.7 & 25.3 & 45.2 & 27.7 & 41.0 & 44.5 & 29.0 & 34.7 & 30.4 & 32.5  \\
                           & Decom (gen.) + I  & 25.6 & 32.2 & 35.9 & 30.6 & 25.1 & 45.1 & 27.7 & 41.0 & 43.7 & 28.6 & 34.6 & 30.3 & 32.5  \\ 
\midrule
\multicolumn{2}{c}{language}                   & ru   & ta   & te   & th   & tl   & tr   & uk   & ur   & vi   & wo   & yo   & zh   & avg.  \\ 
\midrule
\multirow{3}{*}{zero-shot} & Iter              & 6.8  & 5.0  & 5.1  & 6.8  & 3.9  & 9.0  & 5.2  & 6.6  & 4.2  & 1.4  & 7.2  & 7.6  & 7.4   \\
                           & Decom (prob.)     & 26.1 & 15.0 & 7.9  & 8.7  & 7.8  & 15.5 & 23.7 & 8.1  & 14.4 & 11.0 & 18.9 & 21.7 & 17.6  \\
                           & Decom (gen.)      & 27.9 & 20.7 & 12.8 & 2.7  & 1.9  & 17.4 & 28.1 & 12.8 & 25.7 & 21.1 & 28.3 & 26.0 & 20.6  \\ 
\midrule
\multirow{5}{*}{few-shot}  & Iter              & 20.3 & 24.3 & 47.0 & 3.1  & 22.5 & 20.9 & 20.9 & 15.5 & 18.3 & 16.5 & 16.9 & 20.7 & 18.8  \\
                           & Decom (prob.)     & 41.9 & 36.5 & 48.2 & 25.0 & 41.9 & 37.9 & 39.6 & 26.2 & 26.9 & 34.1 & 39.2 & 40.8 & 36.2  \\
                           & Decom (gen.)      & 36.8 & 33.5 & 41.7 & 23.1 & 41.9 & 36.4 & 37.0 & 24.7 & 24.5 & 33.2 & 36.5 & 35.7 & 33.2  \\
                           & Decom (prob.) + I & 37.0 & 34.1 & 39.0 & 13.7 & 57.8 & 38.0 & 35.8 & 26.4 & 34.0 & 30.3 & 33.3 & 32.8 & 32.9  \\
                           & Decom (gen.) + I  & 36.9 & 33.9 & 38.8 & 13.6 & 57.8 & 38.0 & 35.4 & 26.4 & 33.9 & 30.3 & 33.3 & 32.6 & 32.7  \\
\bottomrule
\end{tabular}}
\caption{Full results on BLOOMZ 7b.}
\tablabel{Bloomz-7b1_results}
\end{table*}

\begin{table*}[ht]
\centering
\scalebox{0.75}{
\begin{tabular}{lllllllllllllll} 
\toprule
\multicolumn{2}{c}{language}                 & en   & af   & ar   & bg   & de   & el   & es   & et   & eu   & fa   & fi   & fr   & he    \\ 
\midrule
zero-shot                 & Decom (gen.)     & 47.6 & 45.7 & 37.8 & 48.9 & 48.9 & 45.8 & 40.0 & 45.3 & 41.5 & 44.2 & 46.8 & 42.6 & 42.6  \\ 
\midrule
\multirow{2}{*}{few-shot} & Decom (gen.)     & 49.0 & 41.0 & 16.2 & 37.6 & 43.9 & 31.0 & 37.2 & 34.8 & 33.9 & 33.4 & 32.1 & 38.5 & 34.1  \\
                          & Decom (gen.) + I & 57.3 & 51.9 & 27.4 & 47.2 & 55.4 & 40.1 & 50.1 & 41.2 & 43.6 & 48.1 & 42.4 & 49.9 & 45.6  \\ 
\midrule
\multicolumn{2}{c}{language}                 & hi   & hu   & id   & it   & ja   & kk   & ko   & lt   & mr   & nl   & pl   & pt   & ro    \\ 
\midrule
zero-shot                 & Decom (gen.)     & 40.6 & 38.7 & 39.3 & 39.3 & 32.9 & 46.1 & 29.2 & 47.4 & 47.5 & 42.8 & 46.1 & 40.6 & 49.4  \\ 
\midrule
\multirow{2}{*}{few-shot} & Decom (gen.)     & 23.8 & 33.5 & 39.9 & 36.5 & 14.3 & 32.4 & 17.7 & 37.5 & 34.9 & 42.7 & 36.1 & 37.1 & 35.6  \\
                          & Decom (gen.) + I & 44.7 & 36.2 & 51.9 & 45.7 & 44.6 & 45.7 & 26.7 & 45.7 & 48.8 & 55.3 & 46.2 & 48.9 & 51.5  \\ 
\midrule
\multicolumn{2}{c}{language}                 & ru   & ta   & te   & th   & tl   & tr   & uk   & ur   & vi   & wo   & yo   & zh   & avg.  \\ 
\midrule
zero-shot                 & Decom (gen.)     & 45.9 & 39.4 & 51.3 & 47.1 & 59.3 & 46.9 & 47.4 & 37.9 & 48.4 & 22.3 & 37.5 & 42.8 & 43.1  \\ 
\midrule
\multirow{2}{*}{few-shot} & Decom (gen.)     & 33.5 & 28.1 & 50.9 & 21.9 & 65.7 & 34.7 & 31.2 & 17.7 & 33.9 & 10.5 & 22.4 & 17.2 & 32.5  \\
                          & Decom (gen.) + I & 43.8 & 38.0 & 55.3 & 46.6 & 70.5 & 46.0 & 41.5 & 36.0 & 49.0 & 19.8 & 38.6 & 34.5 & 44.7  \\
\bottomrule
\end{tabular}}
\caption{Full results on mTk 13b.}
\tablabel{mtk-11b_results}
\end{table*}

\begin{table*}[]
    \centering
    \scalebox{0.8}{
    \begin{tabular}{c|ccccccccc}
    \toprule
 $k$   & en   & de   & el   & fa   & hi   & nl   & ru   & zh   & avg. \\
    \midrule
0  & 63.6 & 68.6 & 62.7 & 68.5 & 37.1 & 69.4 & 74.4 & 59.2 & 62.9 \\
1  & 72.4 & 76.1 & 69.2 & 65.9 & 62.5 & 75.1 & 69.4 & 60.2 & 68.9 \\
3  & 77.0 & 75.8 & 70.9 & 70.7 & 69.7 & 71.5 & 75.2 & 70.1 & 72.6 \\
5  & 84.4 & 80.9 & 76.7 & 75.3 & 70.9 & 80.4 & 81.8 & 65.9 & 77.0 \\
7  & 86.2 & 79.8 & 76.5 & 79.3 & 71.4 & 78.8 & 83.2 & 68.3 & 77.9 \\
9  & 85.4 & 78.5 & 72.8 & 78.2 & 73.1 & 77.6 & 81.8 & 66.4 & 76.7 \\
11 & 84.9 & 80.2 & 73.9 & 78.6 & 72.9 & 81.3 & 82.3 & 65.7 & 77.5 \\
13 & 84.1 & 78.7 & 73.1 & 77.9 & 72.6 & 79.1 & 80.7 & 65.4 & 76.5 \\
15 & 86.2 & 80.0 & 73.6 & 71.6 & 73.7 & 82.7 & 83.4 & 67.9 & 77.4 \\
17 & 85.8 & 78.3 & 72.9 & 70.1 & 73.0 & 80.9 & 80.8 & 66.6 & 76.1 \\
\bottomrule
    \end{tabular}}
    \caption{Full results of few-shot ablation study.}
    \tablabel{few_shot_tab}
\end{table*}

\end{CJK}
\end{document}